\pdfoutput=1
\documentclass[11pt,hyperref]{article}
\usepackage{graphicx}
\usepackage{acl}
\usepackage{arydshln}
\usepackage{times}
\usepackage{latexsym}

\usepackage{colortbl}

\usepackage{times}
\usepackage{graphicx}
\usepackage{latexsym}
\usepackage{natbib}
\usepackage{amsmath}
\usepackage{amssymb}
\usepackage{mathtools}
\usepackage{anyfontsize}
\usepackage{booktabs}
\usepackage{csquotes}
\usepackage{multirow}
\usepackage{multicol}
\usepackage{array}
\usepackage{parskip}
\usepackage{setspace}
\usepackage{tikz}

\usepackage[T1]{fontenc}
\usepackage[utf8]{inputenc}

\usepackage{microtype}
\usepackage{url}
\usepackage{graphicx}
\usepackage{booktabs}
\newcommand*\rot{\rotatebox[origin=c]{90}}

\newcommand{\BASIL}{\textsc{BASIL}}
\newcommand{\BANC}{\textsc{BANC}}
\newcommand{\EBTA}{\textsc{EBTA}}
\newcommand{\ABTA}{\textsc{ABTA}}

\title{Target-Aware Contextual Political Bias Detection in News}
\author{Iffat Maab \qquad Edison Marrese-Taylor \qquad Yutaka Matsuo\\
Graduate School of Engineering\\
The University of Tokyo\\
{\tt iffatmaab09@gmail.com}\\ {\tt\{emarrese, matsuo\}@weblab.t.u-tokyo.ac.jp} \\}

\author{Iffat Maab\textsuperscript{1}, Edison Marrese-Taylor\textsuperscript{1,2}, Yutaka Matsuo\textsuperscript{1}\\
\textsuperscript{1}The University of Tokyo\\
\textsuperscript{2}National Institute of Advanced Industrial Science and Technology\\
{\tt \{iffatmaab, emarrese, matsuo\}@weblab.t.u-tokyo.ac.jp}
}

\begin{document}
\maketitle
\begin{abstract}
Media bias detection requires comprehensive integration of information derived from multiple news sources. Sentence-level political bias detection in news is no exception, and has proven to be a challenging task that requires an understanding of bias in consideration of the context. Inspired by the fact that humans exhibit varying degrees of writing styles, resulting in a diverse range of statements with different local and global contexts, previous work in media bias detection has proposed augmentation techniques to exploit this fact. Despite their success, we observe that these techniques introduce noise by over-generalizing bias context boundaries, which hinders performance. To alleviate this issue, we propose techniques to more carefully search for context using a bias-sensitive, target-aware approach for data augmentation. Comprehensive experiments on the well-known BASIL dataset show that when combined with pre-trained models such as BERT, our augmentation techniques lead to state-of-the-art results. Our approach outperforms previous methods significantly, obtaining an F1-score of 58.15 over state-of-the-art bias detection task. 

% showing that adequately sampling context () is critical for improving bias detection. 
% , and to quantify the efficacy of our approach we perform exhaustive ablation studies on some bias tasks.

% We carefully gather different contexts models of state-of-the-art to unveil inconsistent terminologies and provide a thorough comparative evaluation of our own approach, and discuss novel contributions.  

\end{abstract}

\section{Introduction}

News media companies publish thousands of articles every day. While we generally regard these articles as containing factual, true information, studies have shown that various kinds of bias exist in news \cite{fan-etal-2019-plain, lim2020annotating, gentzkow2015media, prat2013political}. Further studies have studied the effects that these biases have on readers, particularly in voting. A study by \citet{groseclose2005measure} suggests that indeed media has a sizable political impact on voting, where for example \citet{dellavigna2007fox} found significant effect of exposure to Fox News in increased turnout to the polls.

Clearly, biased media have the potential to sway readers in potentially detrimental paths. Therefore, it is crucial to unveil the true nature of media bias. Furthermore, as all journalism contains narratives \cite{unesco2018journalism}, given its role on transforming individual and public opinion, we consider it is worth measuring and understanding the political bias phenomenon. We think bias detection is important as a proxy or mechanism to assess the quality of information in news media. As stated by \citet{unesco2018journalism}, there is no problem with the existence of narratives in substandard journalism, rather poor professionalism.

Bias in news from different aspects has been studied in the past, where for example \citet{chen2018learning} and \citet{arapakis2016linguistic} created news quality corpus of 561 articles and study how various news constituents characterize the quality of editorial articles. While these works are highly relevant to the bias problem, they did not specifically or directly target at the issue.

Foundational work in political bias was performed by \citet{fan-etal-2019-plain}, who released a human-annotated dataset named Bias Annotation Spans on the Informational Level (BASIL), containing 300 fine-grained bias annotations. Concretely, political bias is identified at the sentence-level, where spans are annotated and a target (the main entity) is identified, in addition to a few other labels. Significantly, BASIL stands as the first dataset to be annotated with different types of bias. \textbf{Informational} bias, which depends broadly on the context of the sentence \cite{multi-level-context} and arises from manipulation of information or selective presentation of content in a factual way, e.g., use of quotes, to evoke specific reader's emotions towards news entities \cite{fan-etal-2019-plain,van-den-berg-markert-2020-context}, and \textbf{lexical} bias, which  stems from the choice of specific words or linguistic phrases that influence the interpretation of a subject, and perpetuate the understanding of information \cite{recasens-etal-2013-linguistic,iyyer-etal-2014-political, Hube_2019} are present in BASIL. To the best of our knowledge, BASIL is the first dataset that annotates informational bias together with specific targets.

% Out of the total 7,977 BASIL sentences, 1,249, 478, 6,250 contains informational bias, lexical bias, and no bias sentences, respectively. In the 300 articles of BASIL corpus, event context consists of the triplets of 100 articles from three different news sources namely Fox News (FOX), Huffington Post (HPO), and New York Times (NYT), each documenting the same event.

With the release of BASIL, work on political bias detection has mostly focused on informational bias, with a strong emphasis on informational context within and across news media articles, as informational bias is highly content-dependent. In the seminal work, \citet{van-den-berg-markert-2020-context} feed the whole document/article as context for sentence-level bias classification. Though this approach worked relatively well in practice, using long documents in this context brings considerable noise, redundancy and can increase vocabulary size, which can ultimately decrease the performance of the classifier as evidenced by previous work \cite{akhter2020document, guo2022modeling}. Moreover, as shown by \citet{chen2020detecting}, detecting bias at article level remains even more challenging and difficult task.

% who found that models suffer from complex and computationally expensive tasks when trained on large text documents.
In light of this issue, several works have recently focused on introducing more specific contextual information to perform classification \cite{cohan2019pretrained, van-den-berg-markert-2020-context, guo2022modeling}, for example by mixing contexts of informational and lexical bias at both the article-level (entire article encompassing target sentence) and event-level (triplet of articles discussing the same event). 

While the aforementioned approaches have resulted in improved performance, we think their applicability is limited. On one hand, articles in BASIL have no overall bias label, instead each sentence is labeled as evidence of a certain kind of bias or as a neutral statement, suggesting that these should be treated separately when detecting different kinds of bias. Previous studies \cite{rao2018lstm,tripathy2017document} have already shown that on document-level classification, paragraphs can belong to multiple categories, which \citet{chen2020detecting}, also observed on BASIL, where paragraphs belong to either informational bias, lexical bias or no bias spans. Furthermore, as highlighted by \citet{chen2020detecting}, by mixing contexts of informational and lexical bias, it becomes difficult for the model to distinguish and predict different type of bias, which may result in lower model performance. 

% introduces the problem of using more generalized contexts by mixing contexts of two completely different bias i.e., informational and lexical in training bias classification tasks, 

% introduces the problem of using more generalized contexts by mixing contexts of two completely different bias i.e., informational and lexical in training bias classification tasks, 

\begin{table}[t]
    \centering
    \footnotesize
        \scalebox{0.8}{
            \begin{tabular}{l@{\hspace{0.15cm}} c@{\hspace{0.15cm}} c@{\hspace{0.15cm}} p{5cm} c@{\hspace{0.15cm}} } % <-- Alignments: 1st column left, 2nd middle and 3rd right, with vertical lines in between
                \toprule
                %\rowcolor{gray!25}
                \textbf{Source} & \textbf{Target} &  \textbf{Index} & \textbf{Sentence} & \textbf{Bias} \\
                \midrule
                \rowcolor{lightgray!20}
                % {\parbox{1\linewidth}{\vspace{1cm} something something}
                FOX & & 0 &  President Obama health care plan treats the treasured entitlement like a piggy bank, while the Romney-Ryan plan preserves it. & Inf\\
                \rowcolor{lightgray!20}
                HPO & & 4 & If any person in this entire debate has blood on their hands in regard to Medicare, it's Barack Obama.  & Inf \\
                \rowcolor{lightgray!20}
                NYT & \multirow{-3}{*}{\parbox{0.5cm}{\vspace{-1.3cm}\rot{Obama Campaign}}} & 4 & Now when you need it, Obama has cut \$716 billion from Medicare.  & Inf \\
                \midrule
                \rowcolor{lightgray!50}
                FOX & & 21 & Obama campaign spokeswoman Lis Smith described the new Romney-Ryan ad on the subject as dishonest and hypocritical, considering Ryan's own proposals for Medicare. & Lex\\
                \rowcolor{lightgray!50}
                HPO & & 27 & Senator McCain and Governor Romney have subsequently opposed the savings that the president identified and demagogued the issue, ironically, since Governor Romney's running mate kept them in his budget.  & Lex \\
                \rowcolor{lightgray!50}
                NYT & \multirow{-3}{*}{\parbox{0.5cm}{\vspace{-2cm}\rot{Romney Campaign}}} & 7 & Lis Smith, a spokeswoman for the Obama campaign, said, Mitt Romney's Medicare ad is dishonest and hypocritical. & Lex \\
                %\rowcolor[gray]{0.8} \cellcolor{blue!25} & \multicolumn{2}{|c|}{\cellcolor{blue!25}\textbf{Mixed Color}} \\
                \bottomrule
            \end{tabular}
    }
    \caption{Bias sentences extracted from event 0 of BASIL with three news media sources, FOX (0fox; source:fox, event:0), HPO (0hpo; source:hpo, event:0), and NYT (0nyt; source:nyt, event:0), showing a single event can exhibit similar targets and bias types to manifest event-based target aware context.}
    \label{tab:example-basil}
\end{table}

In light of this issue, in this work, we provide a framework to generate more consistent and similar bias contexts to improve performance. As shown in Table \ref{tab:example-basil}, each instance of annotated bias span also identifies the ``target'', i.e., the main entity or topic of the sentence that is also annotated in BASIL. Using this information, our key insight is to create event-level contexts that are target-aware and also sensitive to the bias label.

For example, for the target ``Obama Campaign'', sentences from three different news sources are combined to form a single contextual example for informational bias classification, as highlighted in light gray. A similar procedure is applied for ``Romney Campaign'', where sentences are concatenated to form an example for lexical bias classification, highlighted in dark gray. Inspired by ideas from modeling context in informational bias detection \cite{van-den-berg-markert-2020-context, chen2020detecting, guo2022modeling}, our approach is able to augment examples with richer contexts and less noise, and follows previous work in determining that the detection of lexical bias should hold equal importance as informational bias \cite{zhou2020towards,8946281,maab2023effective}. 

Following recent work \cite{maab2023effective}, we tackle a variety of bias detection tasks including INF/OTH and INF/LEX using data from BASIL. Through extensive experimentation, we demonstrate the effectiveness of our approach by obtaining state-of-the-art performance on all of our studied tasks. In addition, our holistic view on bias enables us to unveil inconsistent terminologies used for contextual information of BASIL, therefore we gather such contexts to improve clarity and uniformity, and to avoid previous work problems as indicated in our comparison with the state-of-the-art.

\section{Related Work}
Media bias has been scrutinized often with nuanced variations and under different contexts through diverse terminologies. Misinformation detection based on linguistic driven approaches are exposed by novel approaches \cite{pan2018content, perez2017automatic}. Powerful players in news media advance their interests by devoting plentiful resources to facilitate controlled communication in politics \cite{entman2007framing}, therefore ideology prediction and trustworthiness of news media draws attention \cite{baly2019multi}, while \citet{hamborg2019automated} highlighted that distinctive contributions can be made by computer scientists to study bias. 

\citet{kulkarni2018multi} proposed an attention-based model to capture high-level contexts of news articles including title, link structure, and news information using both textual content and network structure to leverage cues from multiple views. Contextualized representations of sentences for better understanding of documents are studied using numerous pre-trained language models \cite{cohan2019pretrained, iyyer-etal-2014-political}. Inspired from \cite{cohan2019pretrained}, \citet{van-den-berg-markert-2020-context} work on BASIL to propose several context inclusive models on article and event context, and use three BiLSTMs for encoding FOX, HPO, NYT news documents as triplets. Building upon existing study of \cite{van-den-berg-markert-2020-context}, \cite{guo2022modeling} use multi-level graph attention networks for bias detection by MultiCTX model that use contrastive learning from sentence embeddings to discriminate target sentences. Another recent study on BASIL \cite{lei2022sentence} built distillation models on top of RoBERTa for informational bias classification and explore different types of local and global discourse structures. Similarly, article-level bias classifiers \cite{chen2020detecting} use second order bias features of BASIL to manipulate context information using uncased BERT. Using BASIL, BERT by \citet{devlin2018bert} remain as a major baseline model in majority of previous studies \cite{van-den-berg-markert-2020-context, guo2022modeling}. \citet{chen2020detecting} find that fine-tuned BERT has a strong efficacy and use it to reimplement \cite{fan-etal-2019-plain} results. In light of the findings, our proposed approach also utilize BERT \cite{devlin2018bert}. 

\section{Proposed Approach}
% By taking into account the nested structure of our proposed approach in using contextual information, we divide our work in the following two contexts.

% Due to the highly imbalanced nature of this dataset, it has been classified in variety of ways as identified by \citet{maab2023effective}, with majority focusing on informational bias only. 

% In light of this issue, our approach is sensitive to include context of informational and lexical bias separately on article and event context whilst classifying both. To avoid confusion, we choose to stay with the same naming convention for labelling different bias detection tasks as in \cite{maab2023effective}. Earlier investigations on BASIL has constraints in only utilizing context of informational bias with emphasis on one sole task of classifying informational bias (INF) from other (OTH), combined samples of lexical bias and no bias sentences. 

\subsection{Bias-Aware Neighborhood Context} 
Previous work has shown that phrases surrounding a sentence annotated with bias can be used as local context to perform bias classification, and that this local context can contribute to the ability of models to identify and label types of bias. However, by ignoring the nature of these sentences, existing approaches that utilize neighborhood context \cite{van-den-berg-markert-2020-context, guo2022modeling} can run into problems by introducing ambiguous content, for example when adding sentences that are annotated with the opposite bias. As shown by \cite{van-den-berg-markert-2020-context}, this can also lead to massive data leakage problems across train and test sets.

% Prior studies train complete article for a single bias sentence with extreme data leaks as the model's lack to discern different forms of bias as indicated by \citet{chen2020detecting}. 

To account for the disparity in how different bias contexts are overlooked in previous work, in this paper, we propose to care for the bias label of neighboring sentences, advancing to generate Bias-Aware Neighborhood Contexts (\textsc{BANC}), and adding neighboring sentences to the model input as long as they have a related bias label. Table \ref{table:bias-aware-neighborhood-context-example} shows an example of how this procedure works. Since, our approach is bias-sensitive, sentences with informational and lexical bias are treated separately. Therefore, for a given target sentence with index 1, the former (index 0) and next (index 2) sentences become neighbor sentences of lexical bias as they exhibit no bias span as highlighted in green. Correspondingly, to generate a BANC for informational bias classification, we combine sentences with indices 2, 3 and 4 as highlighted in blue. Teal (green + blue) color is shown by sentence index 2, since it is common between the two BANC text spans. According to the same principle, for cases where the first sentence of an article has bias, next sentence is checked and combined, whereas in the event where it is last sentence, former sentence gets checked and successively combined.

% In the context of bias classification tasks, we also tend to focus and classify an important yet ignored lexical bias \citet{zhou2020towards} from informational to overcome the challenges of different bias detection tasks of this special dataset.

% Therefore, on article level, our bias-aware neighborhood context begin by merging immediate former and next neighbor sentences of either no bias or a same bias as a target sentence is, to benefit from comprehensive and improved contextualized representation of semantically similar bias sentences. Table \ref{tab:contextinflex} show examples of neighborhood context of informational and lexical bias on NYT news section (18nyt; source: nyt, event: 18) of BASIL dataset where informational and lexical bias sentences are treated separately. 

% Our work on splitting the contextual information into individual bias enables us to reveal the significance of our approach in detecting low resource lexical bias of BASIL and experiment with different task formulations to perform bias sensitivity analysis.

\begin{table}[t]
    \centering
    \footnotesize
    \scalebox{0.85}{
    \begin{tabular}{l@{\hspace{0.15cm}} c@{\hspace{0.2cm}} p{5cm} c@{\hspace{0.15cm}}} % <-- Alignments: 1st column left, 2nd middle and 3rd right, with vertical lines in between
    \toprule
    %\rowcolor{gray!25}
    \textbf{Index} & \textbf{Position} & \textbf{Sentence} & \textbf{Bias}\\
    \midrule
    \rowcolor{green!10}
    0 & Neighbor & Israel and Middle East policy have a tendency of surfacing in presidential politics in rather combustible ways. & -\\
    \rowcolor{green!10}
    1 & Target & And a new advertisement that will run in areas of Florida with large Jewish populations attempts to stoke anxiety over American policies in the region, using a news clip of Prime Minister Benjamin Netanyahu of Israel warning of the risks of a nuclear Iran.  & LEX \\
    %\rowcolor[gray]{0.8} \cellcolor{blue!25} & \multicolumn{2}{|c|}{\cellcolor{blue!25}\textbf{Mixed Color}} \\
    \cellcolor{teal!40}2 & \cellcolor{teal!40} Neighbor & \cellcolor{teal!40}  The fact is that every day that passes, Iran gets closer and closer to nuclear arms, Mr. Netanyahu is shown saying.  & \cellcolor{teal!40}  - \\
    \rowcolor{blue!9}
    3 & Target & For dramatic effect, a soundtrack fit for an episodic drama like Homeland plays as the prime minister continues.   & INF \\
    \rowcolor{blue!9}
    4 & Neighbor & The world tells Israel, Wait. There's still time. & - \\
    5 & Neighbor & And I say wait for what? & - \\
    6 & Neighbor & Wait until when? & - \\
    \bottomrule
    \end{tabular}
    }
    \caption{ An article of New York Times section extracted from BASIL showing bias-aware neighborhood context of informational bias in green and lexical in blue.}
    %Sentence index 2 highlighted by Teal (Green + Blue) is common between the two neighbourhood text spans.
    %Neighbouring sentences have no bias.}
    \label{table:bias-aware-neighborhood-context-example}
\end{table}

% \IM{will delete the following later}
% \IM{thus leading to less efficient training process, since the model has no clue to distinguish different type of bias. Therefore, we believe it is essential to train the model with similar bias context, while also addressing data leakage problems. }

% \IM{Previous augmentation approaches on BASIL for adding contextual information rely on the complete article and event contexts with no knowledge of distinguishing lexical and informational bias.}
% \IM{We identify contexts of informational and lexical bias spans individually and generate}

% \IM{We devise TAT to learn from all possible combinations of sentences having same bias and target entity to aid the model in understanding the influence of bias targets to achieve precise predictions in different bias formulation tasks \cite{maab2023effective}. }

% \IM{Similarly, our approach is also able to highlight the new entities having more of informational or lexical bias for accurate binary classification of informational verses lexical bias (provide some statistical evidence)\IM{topic based ablation study write}. }

\subsection{Target-Aware Context}

% % More concise table
% \begin{table}
%     \centering
%     \tabcolsep=0.11cm
%     \scalebox{0.58}{
%         \begin{tabular}{lllccc}
%         \toprule
%         & & &  & \multicolumn{2}{c}{\textbf{Sentences with}}\\
%         \cmidrule{5-6}
%         {\textbf{Event}} & {\textbf{Source}} & {\textbf{Target}} & \textbf{\# of Bias Sentences} & \textbf{INF Bias} & \textbf{LEX Bias}\\
%         \toprule
    
%         \multirow{5}{*}{18} & \multirow{2}{*}{FOX} &  Benjamin Netanyahu & 1 & 1 & - \\
%         &    &  Barack Obama & 6 & 5 & 1\\
%         \cmidrule{2-6}
%         & \multirow{2}{*}{HPO} & Barack Obama  & 4 & 1 & 3\\
%         &    &  Secure America Now & 3 & 2 & 1\\
%         \cmidrule{2-6}
%          & \multirow{1}{*}{NYT} & Secure America Now  &  3 & 2 & 1 \\
%         \bottomrule
%         \multirow{5}{*}{22} & \multirow{2}{*}{FOX} &  Hillary Clinton  & 5 & 5 & - \\
%             &    &  Barack Obama & 2 & 2 & - \\
%         \cmidrule{2-6}
%         & \multirow{2}{*}{HPO} & Barack Obama & 2 & 2 & - \\
%         &    &  Nancy Pelosi & 1 & 1 & - \\
%         \cmidrule{2-6}
%         & \multirow{1}{*}{NYT} & Hillary Clinton  &  4 & 3 & 1 \\
%         \bottomrule
%         \end{tabular} 
%     }
%     \caption{Targets extracted from BASIL dataset from event 18 and 22 across three news sources including FOX, HPO, and NYT showing total number of bias sentences and its type per each target. Note that some targets have absence in some news sources.}
%     \label{tab:contexttarget}
% \end{table}

% Separate Informational and Lexical Bias
\begin{table}[t]
        \centering
        \tabcolsep=0.15cm
        \scalebox{0.50}{
            \begin{tabular}{llccc@{\hspace{0.4cm}}ccccc}
                \toprule
                \multirow{3}{*}{\textbf{}} & \multirow{3}{*}{\textbf{Target}} &
                \multicolumn{3}{c}{\textbf{Sentences}} & \multicolumn{5}{c}{\textbf{Target-aware examples}} \\
                % \cmidrule{6-9}
                % &&&&&&&& \\
                % &  &  \multicolumn{3}{c}{\textbf{Article-level}} & & & \multirow{2}{*}{\textbf{All}}\\
                &  &  & & & \multicolumn{3}{c}{\textbf{Article-level}} & \textbf{Event-level} & \multirow{2}{*}{\textbf{Total}}\\
                % % \cmidrule(lr){3-5}\cmidrule(lr){6-8}
                &  & \textbf{FOX} & \textbf{HPO} & \textbf{NYT} &  \textbf{FOX} & \textbf{HPO} & \textbf{NYT} & & \\
                \toprule
                \multirow{3}{*}{18} & Benjamin Netanyahu & 1 & - & - & 1 & - & - & - & 1 \\
                & Barack Obama & 5 & 1 & -  & 10 & 1 & - & 5 (fox $\times$ hpo) & 16  \\
                & Secure America Now & - & 2 & 2  & - &  1 & 1 & 4 (hpo $\times$ nyt) & 6  \\
                \midrule
                % \cmidrule{3-10}
                \textbf{Total} &  & & & & \multicolumn{3}{c}{within Art. $=$ 14} & 9 & \textbf{23} \\
                % \cmidrule{1-10}
                \midrule
                \midrule
                \multirow{3}{*}{22} & Hillary Clinton & 5 & - & 3  & 10 & - & 3 & 15 (fox $\times$ nyt) & 28  \\
                & Barack Obama & 2 & 2 & - & 1 & 1 & - & 4 (fox $\times$ hpo) & 6  \\
                & Nancy Pelosi & 1 & - & -  & 1 & - & - & - & 1 \\
                \midrule
                \textbf{Total} & & & & & \multicolumn{3}{c}{within Art. $=$ 16} & 19 & \textbf{35} \\
                \midrule
                \bottomrule
            \end{tabular}
        }
    \caption{Detail of the number of contextualized instances obtained by applying our proposed ABTA and EBTA to a set of the original examples from BASIL, in this case taken from events (E) 18 and 22, for the case of informational bias.}
    % sentnces annotated with informational bias sentences extracted from three media sources, FOX, HPO, and NYT with its associated target for \textbf{article-based} context are combined  using the number of possible combinations (\textit{C(n,2)}) as illustrated. For each given target in \textbf{event-based} target context i.e., context across articles, distinct media article sentences are paired in all possible arrangements by joining each bias sentence of one article with another (fox $\times$ hpo $\times$ nyt), respectively. Keep in view that sentences must come from the same event and for every unique target entity, the procedure gets repeated. The same is true for lexical bias examples for each target, however not shown here to ensure simplicity.}
    \label{table:target-aware-context-example}
    \vspace{-0.5cm}
\end{table}

While our neighboring approach helps identify local context relevant for bias classification, we believe that global context, either at the article or event levels, can also be exploited.  To that end, we note that BASIL contains annotations that also identify the ``target'' of a given sentence where either lexical or informational bias is present. This ``target'' label refers to the main entity or topic of the sentence that is annotated, with some of the most prominent targets in \BASIL{} being entities or people that lie at the core of news reports, such as Donald Trump, Romney Campaign, Secure America Now, among others.

We further note that although the frequency of appearance of a given ``target'' varies substantially, as long as we keep the annotated label constant (e.g., lexical), the context remains the same. This motivated us to gather all surrounding linguistic cues pertaining to a specific target at both the article-level and event-level. Concretely, we create target-aware contextual information by making use of all possible combination spans having the same bias and target, and propose article-based target-aware (\ABTA) and event-based target-aware (\EBTA) contexts, which we explain below.

% and  context with its associated bias from multiple news sources in event-based target contexts as in BASIL. 
% have varying degrees of  interpreted as bias in our context which in essence is true for real-world news settings because preconceived human notions influence how they perceive bias \cite{liu2019detecting}.

% Concretely, we find target-aware contexts using all possible combinations of sentences having the same bias label, i.e., informational and the same target. 

As show in Table \ref{table:target-aware-context-example}, using \ABTA{} context, for instance, the target ``Barack Obama'' which has 5 sentences annotated with informational bias in the FOX article and 1 in HPO, generates all possible combinations of two sentences within FOX giving us 10 contextualized examples, and 1 same example in HPO because this article has only one sentence, respectively. Note that possible combinations of sentences within articles are combined in groups of two only, which we do to emulate the natural distribution of occurrence of sentences with the same bias and same ``target''.

 % target sentences in some BASIL articles are highly sparse and may have limited occurrence of only two sentences at maximum as evident from 

\EBTA{} contexts shown in the ``Event-level'' column in Table \ref{table:target-aware-context-example}, are computed for common targets across articles, for instance, the same target ``Barack Obama'' with informational bias appear across FOX and HPO with 5 and 1 sentences, therefore all unique possible combinations in groups of two generates 5 new contextualized examples across the two aforementioned articles.  Finally, following the example in the table for ``Barack Obama'', the combined contexts of \ABTA{} and \EBTA{} give us a total of $10+1+5=16$ contextualized informational bias examples for a single target.  Note that we repeat this procedure for generating target-aware lexical bias contexts. 

Because of the way in which we combine sentences, it is evident that our approach is significant in providing contextualized examples for infrequent targets as well, therefore also contributing towards mitigating imbalanced bias distribution and skewed nature of ``targets'' as observed in \BASIL{} articles \cite{chen2020detecting}. 

\begin{table}
    \centering
    \tabcolsep=0.13cm
        \scalebox{0.9}{
        \footnotesize
        \begin{tabular}{l@{\hspace{0.1cm}}cl}
            \toprule
            \textbf{Target} & \multicolumn{2}{c}{\textbf{Target Aware Context}}\\
            \cmidrule{2-3}
            & \textbf{Sentences}   & \textbf{Possible Combinations} \\
            \midrule
            {Donald Trump} &    340   & 2767 (Inf: 2386, Lex: 381)  \\
            {Barack Obama} & 119 & 619 (Inf: 479, Lex: 140) \\
            {Barack Obama*} & 156 & 870 (Inf: 705, Lex: 165) \\
            %\multirow{3}{*}{Romney Campaign}
            {Hillary Clinton} & 62 & 327 (Inf: 292, Lex: 35)\\
            {Democratic Lawmakers} & 36 & 119 (Inf: 97, Lex: 22)\\
            {Joe Biden} &  32 & 325 (Inf: 241, Lex: 84)\\
            {Paul Ryan} & 25 & 122 (Inf: 97, Lex: 25)\\
            \bottomrule
        \end{tabular}
    }
    \caption{Most frequent bias targets in \BASIL{} across events and their possible combinations using target-aware context. Barack Obama* includes three similar targets: Barack Obama, Obama's administration, Sasha and Malia Obama with 119, 21, and 16 bias sentences.}
    \label{table:target-aware-statistics} 
\end{table}

Using our target-aware techniques, we are able to get more than triple the amount of training examples for training for lexical bias detection (462 sentences v/s 1,551 contextualized examples), and we observe a fourfold increase of examples for informational bias detection (1,221 sentences v/s 4,987 contextualized examples). Please see Table \ref{table:target-aware-statistics} for a detailed explanation on target-aware context generation for the most frequent ``targets'' in \BASIL. Our separate use of lexical bias contexts guarantees that the model is not relying solely on shallow lexical features of a complete article as in  previous work \cite{van-den-berg-markert-2020-context, guo2022modeling}, and instead looking for cues on the same categories or bias type \cite{rao2018lstm, tripathy2017document} relevant for the task at hand.

Since, prior studies focused solely on informational bias and overlooked other bias spans, we surmise that lexical bias detection is also significant as supported by \citet{maab2023effective}, and provide a more concise and sensitive bias narratives with neighborhood contexts together with target-aware contexts. 

Finally, based on successful results reported by previous work \cite{mikolov2013exploiting,maab2023effective}, we additionally use a  backtranslation approach to generate more data, which we apply to our contextualized samples using Spanish as a pivot language. By incorporating multiple viewpoints in our neighborhood and target-aware contexts, we facilitate our model in providing a broad and inherent semantics of biased targets to manifest variations in bias representations. Our extensive experiments will further demonstrate the impact of proposed context in different training settings.

%  Additionally, the 
% \IM{write more about adding the neighbourhood context with TAT and use of back translation as inspired from last study on BAISL also provide some reference of back translation}
% \IM{Check if you can mention the changed BASIL dataset size in a separate paragraph or under a new heading}
% \IM{Mathematical formulation if any you for showing TAT}

\section{Experimental Setup}
% In this section, we present our setup, baselines, ablation study of proposed data augmentation context techniques, comparison with state-of-the-art, and in the end the most frequent BASIL targets and their impact on bias detection tasks is discussed.

To streamline the comparison with prior work \cite{van-den-berg-markert-2020-context,guo2022modeling}, we use a 10-fold cross-validation setting where bias-aware neighborhood and event-based target aware contexts never appear at the same time in non-overlapping train-val-test split sets of 80-10-10, respectively. Average performance of our model using three seed runs is reported in all our experiments.

 % from 1,221 to 2,442, and 462 to 924, respectively, whereas only lexical bias context examples are backtranslated in INF/LEX task due to the noticeably skewed size of lexical bias spans as compared to informational.

% , where size of examples wi informational context samples doubles from 4,987 to 9,974, and lexical from 1,551 to 3,102, respectively.

For the sentence-level bias detection, we perform two classification tasks: detection of informational bias (INF/OTH) and classification of bias type (INF/LEX). Inspired by \citet{maab2023effective}, for INF/OTH bias task we combine \BANC{}, \EBTA{} and \ABTA{} with backtranslation of both examples labeled as containing either informational or lexical bias. For the INF/LEX, we do this only on lexical bias examples.  
% for the INF/OTH task, we also utilize the backtranslation on our target-aware examples. In the same context, backtranslated examples of only lexical bias are used in

% task for target-aware contexts. 

We refer to the original set of examples in \BASIL{}, without augmentation as ``regular''. We do not perform any augmentation techniques for the testing examples . Furthermore, to examine the effectiveness of our proposed components in ablation studies, regular BASIL examples \cite{fan-etal-2019-plain} are augmented with \BANC and target-aware contexts in fractions of 10\%, 20\%, 30\%, 40\%, 50\%, 100\%, and 100\% with BT (additional backtranslated examples).

% The concept of using contextual information in BASIL is relatively new and been used by \citet{van-den-berg-markert-2020-context,guo2022modeling, lei2022sentence}. 

\paragraph{Baselines} Majority of approaches in previous studies concentrate on deep learning methods for identifying media bias. We compare our work with models that use different kinds of BASIL contexts for sentence-level bias detection to ensure consistent and impartial evaluation.  We consider multiple contextual models that address the detection of informational bias, for example, SSC (Sequential Sentence Classification) \citet{Cohan_2019} and its variant WinSCC (windowed Sequential Sentence Classification) \cite{van-den-berg-markert-2020-context}, RoBERTa, ArtCIM for target sentences within an article, and EvCIM for triplets of articles covering the same event \cite{van-den-berg-markert-2020-context,guo2022modeling}. \citet{guo2022modeling} further proposed MultiCTX model and reproduce the results using WinSCC and EvCIM for informational bias detection. We also compare against the fine-tuned RoBERTa model \cite{lei2022sentence}, as well as BERT \cite{maab2023effective, chen2020detecting, devlin2018bert, fan-etal-2019-plain}.

\paragraph{Implementation Details} We use the PyTorch to implement our models, borrowing from HuggingFace \cite{face2021ai}, our classifiers are based on BERT-base \cite{devlin2018bert}, and all our models are trained with $5 \times 10^{-5}$ as learning rate, 32 as batch size, and 15 as a maximum epoch count. We utilize a server with an NVIDIA V-100 GPU for our experiments.

\begin{figure}[t]
    \includegraphics[width=1\linewidth]{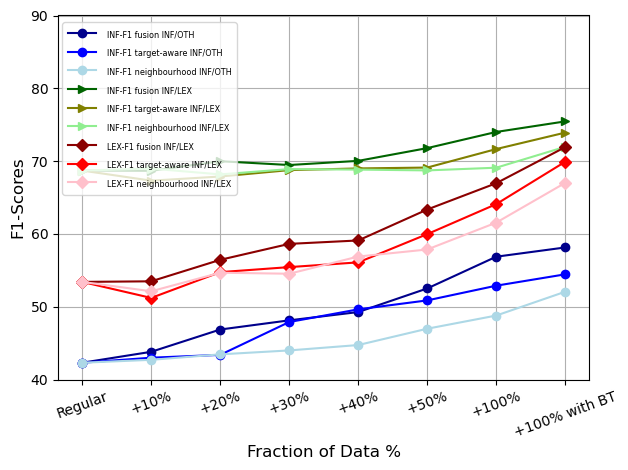}
    \caption{Plot showings various fractions of augmented contexts using \BANC{} (neighbourhood), \ABTA{} \& \EBTA{} (target-aware), and integration of multiple contexts (fusion = BERT + \BANC{} + \ABTA{} + \EBTA{}) to examine the effect of INF-F1 score (blue) in INF/OTH task, and INF-F1 (green) \& LEX-F1 (red) in INF/LEX task.}
    \label{111}
\end{figure}

\begin{table}
    \centering
    \tabcolsep=0.11cm
    \scalebox{0.68}{
    \begin{tabular}{ccccccccc}
        \toprule
        & &  & \multicolumn{3}{c}{\large\textsc{Inf/ Oth}} & \multicolumn{3}{c}{\large\textsc{Inf/ Lex}} \\
        \cmidrule(lr){4-6}\cmidrule(lr){7-9} 
        & & & & \multicolumn{2}{c}{\textbf{F1 Score}} & & \multicolumn{2}{c}{\textbf{F1 Score}}\\
        \cmidrule{5-6}\cmidrule{8-9}
        \textbf{BANC} & \shortstack{\bf ABTA+\\\bf EBTA} & \textbf{\%}   & \textbf{Acc.} & \textbf{INF} & \textbf{OTH} & \textbf{Acc.} & \textbf{INF} & \textbf{LEX} \\
        \midrule
        - & - & Regular             & 80.14 & 42.32	& 87.11 & 76.69	& 68.71 &	53.42\\
        \cmidrule{1-9}
        \checkmark & - & + 10\%      & 80.56 & 42.72 & 86.41 & 77.87 & 68.98 & 52.13 \\
        % \checkmark & - &  + 20\%     & 81.12 & 43.47 & 87.57 & 78.45 & 68.18 & 54.68 \\ 
        \checkmark & - &  + 30\%     & 82.35 & 44.01 & 89.32 & 79.87 & 68.91 & 54.54\\
        % \checkmark & - &  + 40\%     & 82.59 & 44.75 & 89.73 & 81.06 & 68.84 & 56.90\\
        \checkmark & - &  + 50\%     & 83.00 & 46.99 & 89.88 & 80.54 & 68.73 & 57.87\\
        \checkmark & - &  + 100\%    & 83.72 & 48.90 & 90.77 & 79.97 & 69.10 & 61.56\\
        \checkmark & - &  + BT       & 85.31 & 52.07 & 89.97 & 82.21 & 71.97 & 67.02\\
        \cmidrule{1-9}
        - & \checkmark &  + 10\%     & 80.05 & 43.01 & 83.28 & 83.21 & 67.32 & 51.22\\
        % -  & \checkmark &  + 20\%    & 81.06 & 43.40 & 84.06 & 81.99 & 67.89 & 54.76\\
        -  & \checkmark &  + 30\%    & 81.44 & 47.91 & 88.22 & 80.30 & 68.77 & 55.45\\
        % -  & \checkmark &  + 40\%    & 81.90 & 49.65 & 89.66 & 82.28 & 69.00 & 56.10\\
        -  & \checkmark &  + 50\%    & 82.36 & 50.88 & 90.42 & 82.89 & 69.12 & 59.98\\
        - & \checkmark &  + 100\%    & 84.36 & 52.91 & 89.07 & 83.30 & 71.66 &	64.10\\
        - & \checkmark &  + BT       & 86.05 & 54.46 & 91.53 & 86.67 & 73.92 & 69.92\\
        \cmidrule{1-9}
        \checkmark & \checkmark &  + 10\%   & 81.13 & 43.82 & 84.17 & 78.56 & 68.67 & 53.50\\
        % \checkmark & \checkmark &  + 20\%   & 83.12 & 46.88 & 85.72 & 79.80	& 69.99	& 56.45\\
        \checkmark & \checkmark &  + 30\%   & 84.72 & 48.14 & 88.06 & 81.23	& 69.47	& 58.64\\
        % \checkmark & \checkmark &  + 40\%   & 84.48 & 49.28 & 91.00 & 80.96	& 70.04	& 59.11\\
        \checkmark & \checkmark &  + 50\%   & 85.51 & 52.52 & 89.06 & 82.60	& 71.79	& 63.36\\
        \checkmark & \checkmark &  + 100\%  & 84.90 & 56.88 & 90.17 & 83.36	& 74.01	& 66.97\\
        \checkmark & \checkmark &  + BT     & 86.40 & 58.15 & 91.88 & 84.77	& 75.46 & 71.93\\
        \bottomrule
        \end{tabular}
    }
    \caption{Results of our ablation studies, in terms of accuracy and micro-F1 scores, when varying the amount (as percentage) of contextualized examples obtained with \textsc{ABTA} and \textsc{EBTA} that are added to the training data, where BT stands for the backtranslation augmentation approach from \cite{maab2023effective}.}
     % and  ABTA  added to araccuracy and  scores for each percentage of data used for training when data augmentation is used for context inclusion within articles named 'neighbourhood', across articles named 'triplets', and combination of both neighbourhood and triplets using BERT for informational bias detection.
    \label{table:resultsinfandothers}
\end{table}

\section{Results}
% In this Section, we describe our tasks and any task-specific model changes (see ablation studies). 
% In this section, using extensive experiments we provide evidence that our proposed components BANC and target-aware context are essential, and play a significant role in improving the model performance on all bias tasks.

\subsection{Ablation Study}

To show the effectiveness of our proposed techniques, we rely on both INF/OTH and INF/LEX tasks. For \BANC, (\ABTA{} and \EBTA{}), we vary the percentage of augmented data that is added to the training, and compare against the ``regular'' setting. Table \ref{table:resultsinfandothers} and Figure \ref{111} shows a summary of our obtained results. Overall, we observe that with the increase in size of context-augmented samples for both neighborhood and target-aware context, the model yields improvements in F1-scores and accuracy of both bias tasks. Furthermore, we see that by only using \BANC{}, we can achieve substantial performance improvements, regardless of the fact that this technique neglects event information. 

Owing to the fact that target-aware context contains comprehensive data augmentation contexts (\ABTA{} \& \EBTA{}), an elevated performance in INF-F1 scores in INF/OTH and INF/LEX tasks is observed, as shown in Figure \ref{111} with blue and olive lines, over \BANC{} in light blue and light green lines, respectively. Higher percentage of context achieves higher performance, for instance, when 100\% context of (\ABTA{} \& \EBTA{}) is utilized, INF/OTH task shows INF-F1 score of 52.91 against 42.32 of regular, and INF/LEX task shows INF-F1 score of 71.66 against 68.71 of regular, respectively. In INF/LEX task, the rise of LEX-F1 scores highlighted in red are more prominent than INF-F1 highlighted in green  after 50\% context, because number of lexical bias contexts are partially comparable to informational bias contexts, whereas in INF/OTH task the informational bias contexts are still reasonably lower than OTH (non bias + lexical samples). Similarly, since backtranslation is only performed on lexical bias contexts in INF/LEX task, LEX-F1 scores are more amplified than INF F1-scores. 
%In 100\% of bias aware neighbourhood context, there are a total of 1221 neighbourhood informational bias context samples together with regular informational bias examples which are essentially the same size as neighbourhood context samples are. The same is true for lexical bias neighbourhood contexts. 

When we combine our neighborhood augmentation technique (\BANC) and target-aware article-based and event-based contexts (\ABTA{} and \EBTA{}) as our final model, it is observed that the performance begin excelling against the regular even when 40\% of the combined context-augmented examples are fed to the model. Our results further demonstrate the effectiveness of backtranslation-based augmentation technique on \BASIL{}, following the findings of \citet{maab2023effective}, and showing that this technique can be combined with our proposed components to attain further performance improvements. 

 % also incorporate multi-context learning named  by combining 

% to see the total impact of our local (article-based) and global (event-based) bias sensitive contexts in learning bias detection tasks. 
% Shown by the dark green, dark red, and dark blue lines using fusion, it is observed that

% Please see Table \ref{table:resultsinfandothers} for detailed study of our experiments.

\subsection{Comparison with Prior Work}

\begin{table}[t]
    \centering
    % \footnotesize
    \scalebox{0.63}{
    \begin{tabular}{l@{\hspace{0.2cm}} c@{\hspace{0.2cm}} c@{\hspace{0.2cm}} c@{\hspace{0.2cm}} c@{\hspace{0.2cm}}}
    \toprule
    \multirow{2}{*}{\bf Model} & \multicolumn{4}{c}{\bf INF / OTH} \\
    \cmidrule(lr){2-5}
    & \bf {Acc.}      & \bf{P} & \bf{R}& \bf  INF F1     \\ 
    \midrule
    \textbf{Neighborhood Context} & \\
    SSC-5 \cite{van-den-berg-markert-2020-context}      & -     & 41.90   & 36.16      & 38.19            \\
    SSC-10 \cite{van-den-berg-markert-2020-context}     & -    & 43.84   & 34.88     & 38.22            \\
    WinSSC-5 \cite{van-den-berg-markert-2020-context}   & -     & 42.28   & 36.94     & 38.67            \\
    WinSSC-10 \cite{van-den-berg-markert-2020-context}  & -     & 43.20   & 35.12     & 37.44\\
    RoBERTa \cite{van-den-berg-markert-2020-context}    & -     & 43.12   & 41.29      & 42.16 \\
    MultiCTX \cite{guo2022modeling}                     & -     & 47.18   & 44.01 & 45.53 \\  
    BERT + BT \cite{maab2023effective}                  & 83.86 & \textbf{51.22} & 46.32 & 50.70 \\
    BERT + \BANC{} (ours)                                  &  83.72 &   49.07 &   45.32      &  48.90  \\
    BERT + \BANC{} + BT (ours)                             & \bf 85.31 &  50.08 &   \bf 48.12      & \bf 52.07  \\
    \midrule
    \midrule
    \textbf{Article Context }\\
    WinSSC \cite{guo2022modeling}                       & -     & 41.47   & 34.37     & 37.58\\ % Text Chunks
    ArtCIM \cite{van-den-berg-markert-2020-context}     & -    & 38.81 &  47.78         & 42.80 \\
    \midrule
    % RoBERTa  \cite{van-den-berg-markert-2020-context}(not known which? & -     & 43.12   & 41.29      & 42.16            \\
    % (not known which? article or event, also same value as direct) & & & & \\
    \textbf{Event Context} \\
    EvCIM \cite{van-den-berg-markert-2020-context}      & -  & 39.72 & 49.60 & 44.10 \\
    EvCIM \cite{guo2022modeling}                        & - & 47.07 & 44.64 & 45.81 \\
    \midrule
    \midrule
    BERT \cite{chen2020detecting}                       & - & 58.62 & 32.08 & 41.46 \\
    RoBERTa \cite{lei2022sentence}                     & - &43.53  & 49.84              & 46.47            \\
    MultiCTX \cite{guo2022modeling}                     & - & 47.78 & 44.50             & 46.08            \\
    BERT + \ABTA{} + \EBTA{} (ours)                           & 84.36 & 52.78  & 47.74 &  52.91  \\
    BERT + \ABTA{} + \EBTA{} BT (ours)                        & \bf 86.05 & \bf 54.10  & \bf 49.82  & \bf 54.46  \\
    BERT + \BANC{} + \ABTA{} + \EBTA{} (ours)                    &  84.90 &   55.60      & \bf 53.93          &  56.88  \\
    BERT + \BANC{} + \ABTA{} + \EBTA{} + BT (ours)               & \bf 86.40 &   \bf 59.22      &  53.12          & \bf 58.15  \\
    \bottomrule
    \end{tabular}
    }
    \caption{Comparison of our approach with previous work, separated by usage of context. We report average results of three runs with different random seeds. In the Table, Acc, P, and R stand for Accuracy, Precision and Recall respectively. BT denotes the augmentation approach from \cite{maab2023effective}, who are also the only authors to report accuracy.}    % . In our experiments the best score obtained on a single run is INF-F1 59.26
     % }
    \label{table:previous_work}
    % \vspace{-0.5cm}
\end{table}

Having established the efficacy of our proposed approach, we now proceed to compare our model with previous studies. Concretely, we can only compare our work against one studied INF/OTH bias task of \BASIL{} using contextual information as indicated by prior work \cite{maab2023effective}, therefore we solely present our work pertaining to this task with state-of-the-art. 

Based on our comprehensive analysis on how prior studies use different contexts on \BASIL{}, we align similar contexts of our proposed method to allow meaningful comparisons as shown in the Table \ref{table:previous_work}, using three corresponding sections.

To compare with previous work where only within article context is used, we concretely utilize our top performing models for comparison, i.e., BERT combined with 100\% \BANC{} (BERT + \BANC{}), and with backtranslation (BERT + \BANC{} + BT). Similarly, prior work using event contexts are compared with our BERT model trained on 100\% target-aware (BERT + \ABTA{} + \EBTA{}), and with backtranslation (BERT + \ABTA{} + \EBTA{} + BT), respectively. Since MultiCTX by \citet{guo2022modeling} uses multi-contrast learning of both article and event contexts, we compare and use our best BERT model with fusion of both proposed context techniques (BERT + \BANC{} + \ABTA{} + \EBTA{}), and with backtranslation (BERT + \BANC{} + \ABTA{} + \EBTA{} + BT), which in essence is our final model. 
%  to draw insightful conclusions.
Based on our results, and supporting findings of our ablation study, both \BANC{} and target-aware (\ABTA{} \& \EBTA) hold significance in our approach, however target-aware contexts contributes more than \BANC{} parallel to previous findings \cite{guo2022modeling}. Our approach outperforms previous work significantly, obtaining an F1-score of 58.15 in INF label.

In summary, our results show that our proposed approach leads to state-of-the-art results, offering compelling empirical evidence suggesting that adding multiple contextual information is effective at recognizing sentence-level informational and lexical bias as a type of misinformation. 

% Therefore, we conclude that bias sensitive subspace and target-aware contexts from a large and diverse set of news media may lead to improved bias classification when compared with previous studies \cite{van-den-berg-markert-2020-context, guo2022modeling}, consequently discussed in following section.

% Through experiments, we demonstrate that our proposed method significantly outperforms the previous work, delivering state-of-the-art results.

\subsection{Role of ``target'' frequency}

%by giving appropriate proportion to different targets according to their presence. 
To confirm the effectiveness of using target-aware (\ABTA{} \& \EBTA{}) contexts, we conduct a study on most frequent bias targets of BASIL, and consequently experiment with BERT \cite{devlin2018bert}, which serves as a baseline model for recent studies \cite{van-den-berg-markert-2020-context, guo2022modeling, chen2020detecting}. From Table \ref{table:target-aware-statistics}, we see that the ``target'' ``Donald Trump'' appears as the most attracted and significant media entity with substantial coverage of informational and lexical bias sentences. Out of 6,538 total target-aware contexts that we create, we found that around 42\% (2,767) of them come from this target. In light of this issue, we are interested in studying the effect of target frequency in the creation of richer context, and propose an ablation analysis to gain insight. 
% in our ablation study, it is natural for us to combine them.

% both positive and negative, though negative spans have high prevalence \cite{fan-etal-2019-plain}. 

We begin by first introducing target-aware contexts of only ``Donald Trump'' in various fractions to the regular setting, again for the INF/LEX task. We compare the contribution of the most frequent target towards performance by testing models trained solely on this data, and compare to models trained on the entire target-aware contexts.

\begin{figure}[t]
    \includegraphics[width=0.5\textwidth]{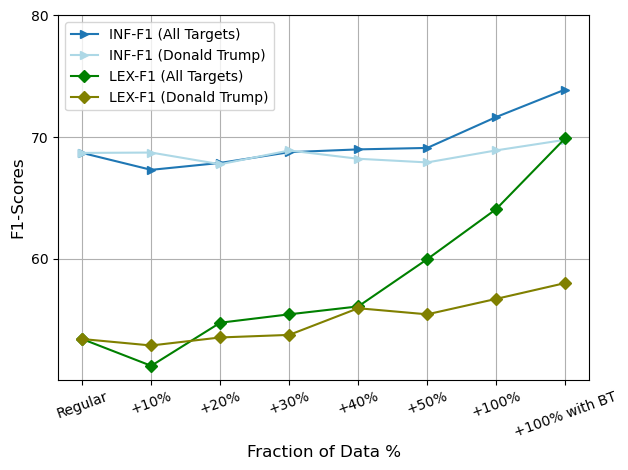}
    \caption{In INF/LEX task, plot showing comparison of performance on most frequent target ''Donald Trump'' v/s. All Targets-aware context (BERT + \ABTA{} + \EBTA{}) using INF/LEX bias task.}
    \label{fig5INFLEX}
\end{figure}

As shown in Figure \ref{fig5INFLEX}, our results are consistent with the performance rise of LEX-F1 scores after 50\% data using all targets, no significant performance change is observed until 50\% of Donald Trump contexts, however there is a gradual rise in F1-scores of INF and LEX when 100\% contextual data is introduced. Increase in LEX F1-score of 56.72 from 53.42 is seen with 100\% Donald Trump context when compared with the regular model. Similarly, due to the fact that no BT is performed on INF bias contexts, the rise of LEX-F1 scores from 56.7 to 58.02 is more prominent in 100\% with BT. In addition to the non-overlapping train-val-test, for this study we carefully choose testing examples so that the majority of targets have an equal \%age in the test set to avoid the problems of overfitting the same target. 

%Parallel to the ablation study performed on all topics in section 5.1, rise of LEX-F1 scores after 50\% data have more prominence because when they are augmented with 10\% fraction increase till 50\% of total 381 Donald Trump lexical contexts, there is little to no change in the total fraction of lexical training data, however when entire 100\% lexical contexts are introduced, the model performance accelerates. 

%Consistent with the performance rise of LEX-F1 scores after 50\% data using all targets is because only lexical examples are augmented in IN

\begin{figure}[t]
    \includegraphics[width=1\linewidth]{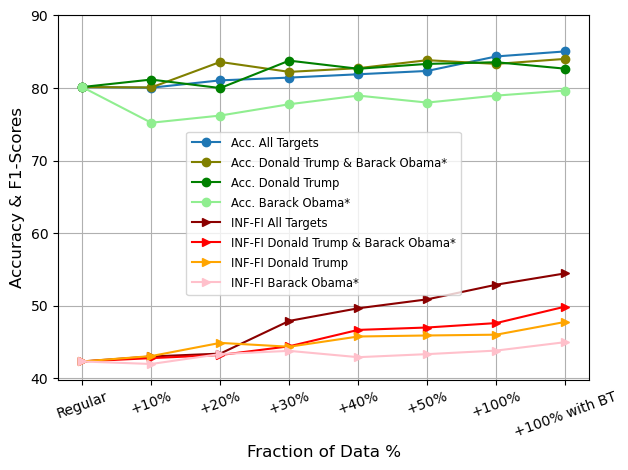}
    \caption{In INF/OTH task, plot showing comparison of F1-scrore and accuracy on target ''Donald Trump'', ''Barack Obama*'', and ''Donald Trump + Barack Obama*'' combined Vs. All Targets-aware context (BERT + \ABTA{} + \EBTA{})}
    \label{fig6INFLEX}
\end{figure}

In addition to Donald Trump in INF/OTH task, we also introduce the second most frequent target ``Barack Obama*'', and the fusion of the two as shown in Figure \ref{fig6INFLEX}. Consistent with our findings, our approach works well for even a single target like ``Donald Trump'' having approximately not far from half target-aware contextualized examples towards total. Following prior work \cite{maab2023effective}, we also provide our model with back translated examples of Donald Trump, hence doubling up context examples from 2,767 to 5,534. Compared to the context free regular model, the best performance of INF F1-score of 46.01 from 42.32 is achieved, whereas through back translation INF F1-score further increased from 46.01 to 47.77 using Donald Trump. The combined context of ``Donald Trump'' and ``Barack Obama*'' is also examined for further confirmation of our proposed target-aware context approach which result in improved performance over a single target ``Donald Trump''. This study confirms the general nature of our approach in detecting different types of bias, since it is not uncommon in real world scenarios to run into similar and parallel target entities as reported by various media outlets \cite{arapakis2016linguistic, lim2020annotating}.

\begin{table*}[t]
    \centering
    \footnotesize
        \scalebox{0.8}{
            \begin{tabular}{l@{\hspace{0.15cm}} c@{\hspace{0.15cm}} c@{\hspace{0.15cm}} c@{\hspace{0.15cm}} p{11.5cm} c@{\hspace{0.15cm}} } % <-- Alignments: 1st column left, 2nd middle and 3rd right, with vertical lines in between
                \toprule
                %\rowcolor{gray!25}
                \textbf{Source} & \textbf{Target} &  \textbf{Local Index} & \textbf{Global Index}& \textbf{Sentence} & \textbf{Bias} \\
                \midrule
                \rowcolor{lightgray!20}
                % {\parbox{1\linewidth}{\vspace{1cm} something something}
                FOX & & 3, 5 & 4 &  Your actions both past and present are incompatible with your duty as Chairman of this Committee, the letter stated. \textbf{We have no faith in your ability to discharge your duties in a manner consistent with your Constitutional responsibility and urge your immediate resignation as Chairman of this Committee.}The letter follows the conclusion of Special Counsel Robert Mueller's Russia probe, which turned up no evidence of collusion between Trump campaign members and Russia during the 2016 presidential election. & Inf\\
                \rowcolor{lightgray!20}
                HPO & & 5 & 6 & It doesn't appear that was any part of [special counsel Robert] Mueller's report. \textbf{In a letter dated Thursday, the GOP committee members accused Schiff of standing at the center of a well-orchestrated media campaign about a possible Trump-Russia connection.}  & Inf \\
                \rowcolor{lightgray!20}
                NYT & \multirow{-3}{*}{\parbox{0.5cm}{\vspace{-0.9cm}\rot{Adam Schiff}}} & 19, 21 & 20 & They say Democrats will stop at nothing to ruin his presidency, and bristle at Democrats accusing them of turning a blind eye to the Russian threat. \textbf{And at the center of their wrath is Mr. Schiff, whose doughy-faced demeanor hardly evokes an attack dog.} The findings of the special counsel conclusively refute your past and present assertions and have exposed you as having abused your position to knowingly promote false information, having damaged the integrity of this committee, and undermined faith in U.S. government institutions, Representative K. Michael Conaway, Republican of Texas, said to Mr. Schiff. & Inf \\
                \midrule
                \rowcolor{lightgray!50} %33 event
                FOX & & - & - & -  & - \\
                \rowcolor{lightgray!50}
                HPO & & 9, 11 & 10 & Previously, she had spoken out against the military's former ''don't ask, don't tell'' policy. \textbf{According to The Hill, Cheney sought to clarify her position after an alleged poll in Wyoming said she "supports abortion and aggressively promotes gay marriage.} Her opposition also puts her at odds with her father, who offered support for gay marriage in 2009. & Lex \\
                \rowcolor{lightgray!50}
                NYT & \multirow{-3}{*}{\parbox{0.5cm}{\vspace{-1.2cm}\rot{Liz Cheney}}} & 7, 9 & 8 & That position deferring to the will of the voters on a state-by-state basis may represent something of a compromise between total support or opposition. \textbf{But it did little to placate her sister.} It's not something to be decided by a show of hands, Mary Cheney wrote. & Lex \\
                %\rowcolor[gray]{0.8} \cellcolor{blue!25} & \multicolumn{2}{|c|}{\cellcolor{blue!25}\textbf{Mixed Color}} \\
                \bottomrule
            \end{tabular}
    }
    \caption{Combined bias sentence examples with three news media sources extracted from event 7 (7fox, 7hpo, 7nyt) for informational bias and event 33 (33fox, 33hpo, 33nyt) for lexical bias showing the influence of local and target-aware global contexts that aids the model in effectively determining bias. In this example we see how sentences in bold, representing bias target sentences with global indexing (event-based), are harmoniously integrated with contextual information from neighboring sentences (local indexing, i.e., preceding and subsequent sentences within the article.)}
    \label{tab:influence-example-basil}
\end{table*}

% \IM{
Furthermore, the incorporation of appropriate context in training samples serves significantly in enhancing the model's performance. To further illustrate this, Table \ref{tab:influence-example-basil} shows two examples of combined local (article-based) and global (event-based) contexts of informational and lexical bias for targets Adam Schiff and Liz Cheney, respectively. It can be seen that meaningful combination of local bias-sensitive contexts and a target-aware context approach in the examples are combined with target sentences which enables the model to detect various types of bias with increased precision, as shown by our results.
% }

\section{Conclusion}
We study a challenging and significant task of detecting misinformation and shed light on bias prevalence in news media. Our work focuses on incorporating bias sensitive (BANC) and target-aware contexts (ABTA \& EBTA) for sentence-level bias detection tasks. Our proposed approach exploits the distinct influence of informational and lexical bias in news media writing styles, emulating the principle of human learning. Our model encompass the process by which individuals acquire new knowledge in real-world settings, i.e., gathering the associated type of bias from common news media targets covering the same event coupled with past experiences, and subsequently utilizing such contexts to make predictions about unfamiliar aspects. 

Our model concretely outperforms classification performance of strong baselines in all bias tasks and we provide statistical significance of our proposed components through extensive experiments. We find that the best performance is achieved when target-aware contexts are combined with BANC, and our methodological stand-point in using small-augmented data of frequent targets suggests that our model is better at recognising bias in mass media. In addition, we conclude its important to keep different bias separately for accurate prediction of bias and we intend to explore other bias features as part of future work. Consequently, future work could also extend contextual information to other misinformation tasks.

\section*{Limitations}
Bias can vary based on human perspectives and existing NLP models have limitations to interpret the subjective nature of bias. Due to the lack of bias representations and annotated media coverage in other languages, our work is based only on English news articles. To the best of our knowledge, BASIL is the only dataset annotated with informational bias, and although our approach provides valuable insights and findings on detecting bias, we provide no evidence to suggest the significance of our findings regarding other contexts surrounding bias or misinformation detection tasks. Similarly, due to the disproportionate number of political ideologies in our dataset, we cannot say for sure if our model will perform equally well for other tasks, and we believe this requires further analysis.

\section*{Ethical Considerations}
In this work, since we highlight some frequent bias targets in political news to propose the significance of our approach, we do not intend to promote media bias entities rather we advocate media literacy and ethical journalism practices. Further, the results we reported in our work highlight deeper understanding of bias contexts, and the need for bias mitigation at various levels of the mass media. 

\section*{Acknowledgements}
The authors wish to express gratitude to the funding organization as this work has been supported by the Mohammed bin Salman Center for Future Science and Technology for Saudi-Japan Vision 2030 at The University of Tokyo (MbSC2030).

% English news articles may also be useful for ana-
% lyzing bias, and require further research analysis to
% verify media bias. Similarly, though our proposed
% approach works well for detecting bias in BASIL,
% we provide no evidence to suggest if this will also
% work on other misinformation-related tasks. The
% same applies for models other than the ones we
% tested in this paper, which though includes a broad
% selection (SVMs, LSTMs and Transformers) is not
% completely comprehensive

% These three news outlets base their stories on the same events, but manage to convey strikingly
% different impressions of what actually transpired. It is such systematic differences in the mapping
% from facts to news reports — that is, differences which tend to sway naive readers to the right or
% left on political issues — that we will call bias.

\newpage
    
\bibliography{target}

% Semi-supervised topic modeling for gender bias discovery in English and Swedish

\clearpage

%\appendix

%\section{Computational Resources}

%We utilize a server with an NVIDIA V-100 GPU for our experiments.
\end{document}